\documentclass[conference]{IEEEtran}
\IEEEoverridecommandlockouts
\usepackage{cite}
\usepackage{amsmath}
\usepackage{amsmath,amssymb,amsfonts}
\usepackage{algorithmic}
\usepackage{graphicx}
\usepackage{textcomp}
\usepackage{xcolor}
\usepackage{hyperref}
\usepackage{listings}

\definecolor{codegreen}{rgb}{0,0.6,0}
\definecolor{codegray}{rgb}{0.5,0.5,0.5}
\definecolor{codepurple}{rgb}{0.58,0,0.82}
\definecolor{backcolour}{rgb}{0.95,0.95,0.92}

\lstdefinestyle{mystyle}{
    backgroundcolor=\color{backcolour},   
    commentstyle=\color{codegreen},
    keywordstyle=\color{magenta},
    numberstyle=\tiny\color{codegray},
    stringstyle=\color{codepurple},
    basicstyle=\ttfamily\footnotesize,
    breakatwhitespace=false,         
    breaklines=true,                 
    captionpos=b,                    
    keepspaces=true,                 
    numbers=left,                    
    numbersep=5pt,                  
    showspaces=false,                
    showstringspaces=false,
    showtabs=false,                  
    tabsize=2
}
\usepackage{eso-pic}
\newcommand\AtPageUpperMyright[1]{\AtPageUpperLeft{
 \put(\LenToUnit{0.5\paperwidth},\LenToUnit{-1cm}){
     \parbox{0.5\textwidth}{\raggedleft\fontsize{9}{11}\selectfont #1}}
 }}
\newcommand{\conf}[1]{
\AddToShipoutPictureBG*{
\AtPageUpperMyright{#1}
}}

\def\BibTeX{{\rm B\kern-.05em{\sc i\kern-.025em b}\kern-.08em
    T\kern-.1667em\lower.7ex\hbox{E}\kern-.125emX}}

\begin{document}

\title{Automated Large-scale Class Scheduling in MiniZinc }

\conf{2020 2nd International Conference on Sustainable Technologies for Industry 4.0 (STI), 19-20 December, Dhaka}

\author{
    \IEEEauthorblockN{
    Md. Mushfiqur Rahman\IEEEauthorrefmark{1},
    Sabah Binte Noor\IEEEauthorrefmark{2},
    Fazlul Hasan Siddiqui\IEEEauthorrefmark{3}}
    \IEEEauthorblockA{
    Islamic University of Science and Technology, Gazipur
    \\mushfiqur11@iut-dhaka.edu\IEEEauthorrefmark{1},
    \\Dhaka University of Engineering and Technology, Gazipur
    \\\{sabah\IEEEauthorrefmark{2}, 
    siddiqui\IEEEauthorrefmark{3}\}@duet.ac.bd
    }
}

\IEEEoverridecommandlockouts
\IEEEpubid{\makebox[\columnwidth]{978-1-7281-9576-6/20/\$31.00~\copyright2020 IEEE \hfill} \hspace{\columnsep}\makebox[\columnwidth]{ }}

\maketitle

\begin{abstract}
Class Scheduling is a highly constrained task. Educational institutes spend a lot of resources, in the form of time and manual computation, to find a satisficing schedule that fulfills all the requirements. A satisficing class schedule accommodates all the students to all their desired courses at convenient timing. The scheduler also needs to take into account the availability of course teachers on the given slots. With the added limitation of available classrooms, the number of solutions satisfying all constraints in this huge search-space, further decreases.

This paper proposes an efficient system to generate class schedules that can fulfill every possible need of a typical university. Though it is primarily a fixed-credit scheduler, it can be adjusted for open-credit systems as well. The model is designed in MiniZinc \cite{nethercote2007minizinc} and solved using various off-the-shelf solvers. The proposed scheduling system can find a balanced schedule for a moderate-sized educational institute \emph{in less than a minute}.

\end{abstract}


\section{Introduction}
Class Scheduling, in general, refers to the task of making a routine or weekly schedule for every relevant entity in an educational institution. A routine for an entity consists of classes in multiple time-slots, spread across all working days. Teachers, students, rooms, and courses are primary entities in a class scheduling task. A class scheduler has to prepare a separate schedule for each entity in the system and make sure that individual schedules are mutually consistent. 

\subsection{Class Scheduling in Universities}
A university as an educational institute, needs to prepare routines for its teachers and students. Usually, universities offer courses to their students in one of two ways - a fixed credit system or an open credit system. Two systems vary in the level of openness of choice for students to choose courses. 

The fixed credit system, also known as the lockstep system, is used in many universities around the globe. Universities following the fixed-credit system, usually create fixed tracks of courses for particular programs in each department. A student, at the time of enrollment, has to choose one of the fixed tracks for their coursework. In some cases, universities allow students to make a few tweaks to their track, for example, choosing a major or minor. These tweaks need to be taken into account while preparing the routine. This entire process makes the class scheduling task a very complicated and highly constrained problem. This paper primarily deals with the fixed credit system.


In this paper, class scheduling is defined as the task of preparing a course timetable for all lectures in a week for a particular program where each lecture is assigned to a classroom and a time-slot. Indivisible groups of students attending all lectures together are treated as homogeneous units. Each such homogeneous unit is termed as a ``section". Each lecture has one or more corresponding sections, and a corresponding faculty or teacher. The scheduler’s task is to ensure that the teacher is available for the lecture at the assigned time and there are no conflicts among the timing of different assigned lectures.

\subsection{Complexity of the Problem}

A large-scale scheduling task creates a huge search space. For $ s $  sections, each attending $ c $ courses of $ h $ credit hours each, the total number of lectures per week becomes $ c*h*s $. For $ n $ time-slots per day and considering a five-day week, there are $ 5*n $ slots in a week. A university with $ r $ classrooms has to assign each lecture to one of $ 5*n*r $ possible combinations of classrooms and time-slots. So, the total search space size becomes $ (c*h*s) * (5*n*r) $. Searching through this huge search-space has a computational complexity of $\mathcal{O}(n^4)$.

Within this huge search space, the number of possible and consistent solutions is only a few. When a teacher is assigned a slot for a lecture, the same slot needs to be assigned to the corresponding section for that course. For that slot, the teacher and the section, must both be available. There also has to be a vacant classroom for that slot. All these constraints make the scheduling task super difficult.

\subsection{Traditional Approaches}
Traditionally, most educational institutes have opted for a manual approach to solve their scheduling problem. However, the complexity of the task makes it difficult for humans to find an optimum solution in reasonable time. A human or a team of humans has to spend hours if not days to prepare a fully consistent routine. 

The alternative of human effort can be achieved through the use of a computerized method. However, considering that the number of consistent solutions is significantly small compared to the vast search space, the task is not solvable by the normal searching algorithms. Non-AI search algorithms, like, breadth-first search, and depth-first search, will need a huge amount of time to obtain a proper solution. As both manual and non-AI approaches fall short, researchers and practitioners have often opted for an AI approach. There are plenty of AI solutions to this problem but none of them have global recognition due to different shortcomings. 

This paper introduces a new AI solution to this problem. This novel solution leverages the MiniZinc modeling language where the entire problem is defined. MiniZinc is a high-level declarative modeling language. It is a simple but expressive tool for constraint satisfaction problems. MiniZinc allows the user to write a solver-independent general model. Additional data is passed in as parameters to the model. At the time of execution, the MiniZinc model is internally converted into FlatZinc\cite{becket2014specification} and solved using one of the many available solvers. This paper uses a variety of solvers, and among them, Chuffed \cite{chu2018chuffed} solver outperformed the rest.

\section{Related Works}
Class scheduling with AI techniques has been a well-studied topic of research for a long time now. Much work has been done in this field. Some uses knowledge-based approaches \cite{lee1995clxpert} and some uses the graphical approaches \cite{dandashi2010graph}. Other approaches, like, the use of Instruction Level Parallelism (ILP) techniques \cite{wasfy2007solving}, are also noteworthy.

Lee et al. \cite{lee1995clxpert} describes a knowledge-based approach to solving the task. Their scheduling system is based on the computation of the desirability map. Information about classes and teachers are collected and passed as the input to the system. The system generates a class schedule that meets the specified requirements and constraints. They use the rule-based language, CLIPS \cite{riley1999clips} and they name their system CLXPERT. Their knowledge-based approach was not new to the scheduling domain and was also previously used by Giarrantano et al. \cite{giarratano1998expert}. Noronha et al. \cite{noronha1991knowledge} also proposes such an approach.

Carter et al. \cite{carter2000comprehensive} describes the course time-tabling system for the University of Waterloo. The paper gives a comprehensive description of the scheduling system for an open-credit university. All the steps, starting from choosing courses to allocating classrooms to each course are elaborately described in this paper. The major difference between their problem statement and ours is, they need to allocate each student to lectures (since each student has the flexibility to choose any of the offered courses) whereas, we can directly assign sections to courses. They consider each student as an individual entity, initially. So the complexity of the problem increases hugely. To solve it, they use ``Homogeneous Sectioning". In this approach, they make small clusters of students who request for the same courses and consider these small clusters as homogeneous sections. So, when the time-table is made, these homogeneous sections, individually, are considered as separate entities. Our approach of using ``sections" as homogeneous units is borrowed from this work.

Sampson et al. also addresses a similar problem in their paper \cite{sampson1995class}. They address a specific scheduling problem -- the class scheduling of Darden Graduate School of Business Administration of the University of Virginia. As the curriculum involves two types of systems -- the fixed credit, and the open credit, they split the task into two portions and solve them using a heuristic approach. They split the work into two sub-problems - enrollment and scheduling. The scheduling task resonates with our work. The core concept of their work is participation maximization. For this, they devised an algorithm to maximize the number of students. The students are asked to give their preferences. The algorithm uses a heuristic approach to maximize student participation. Though this is an efficient way of targeting the problem, this falls short for open-credit systems.

Abramson \cite{abramson1991constructing} solves a school time-table scheduling problem. His approach involves the use of Simulated Annealing. Though the problem we are dealing in this paper is not school time-tabling, but a university class-scheduling is not too different from a school class-scheduling. Abramson's work shows remarkable outcomes. Aycan et al. \cite{aycan2009solving}, had also used the simulated annealing approach. They have defined a set of constraints and used simulated annealing algorithm as the searching algorithm. The constraints they used in their system are similar to what we have used in this paperexcept for a few basic differences.

The heuristic approach has given promising results outside of the class scheduling problem too. For example, in the papers \cite{morton1993heuristic, he2003qos, casanova2000heuristics} heuristic approach has been used to solve different types of search and scheduling problems. The outcomes are remarkable.

\section{System Details}
\begin{figure*}
    \centering
    \includegraphics[width=\linewidth]{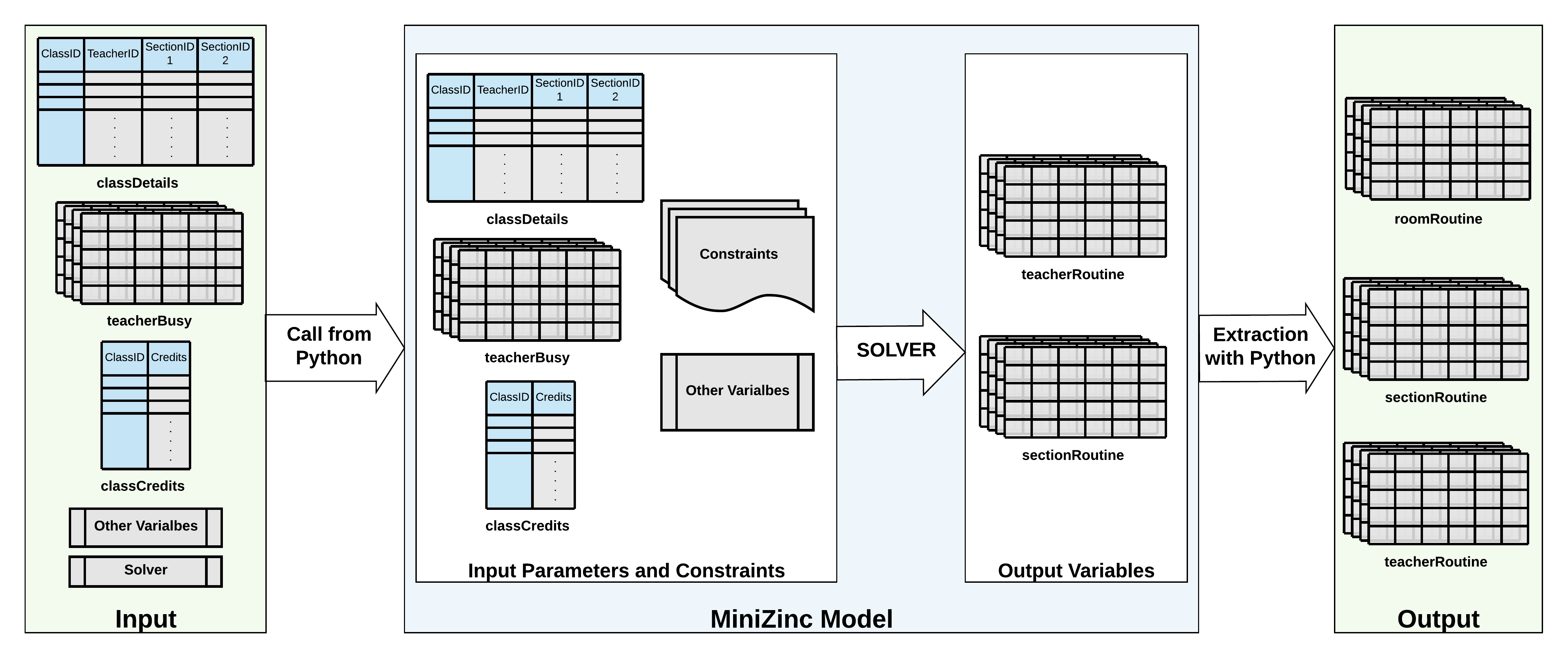}
    \caption{Diagrammatic representation of the developed system}
    \label{fig_model}
\end{figure*}

Fig ~\ref{fig_model} describes the overall system used in this paper. The system uses MiniZinc as the modeling language where the data structures are represented and all the constraints are defined. A designated solver fills data elements according to the constraints. The entire system is wrapped in python for easy I/O operations using python interface.

\subsection{Environment and Data-structures}
MiniZinc is a high-level modeling language, where declaring relevant variables and parameters and their relations in the form of constraints, is enough to generate an ``optimized" or ``satisfiable" solution, with the help of a solver.

The primary task of this system is to generate routines for each entity, i.e. , for each teacher and each section, keeping logical coherence. Teachers and sections are the only concerning entities. Therefore, they need to be represented as variables. This work intentionally avoids other relevant personnel in the system for the ease of representation.

\begin{figure}[htbp!]
    \centering
    \includegraphics[width=\linewidth]{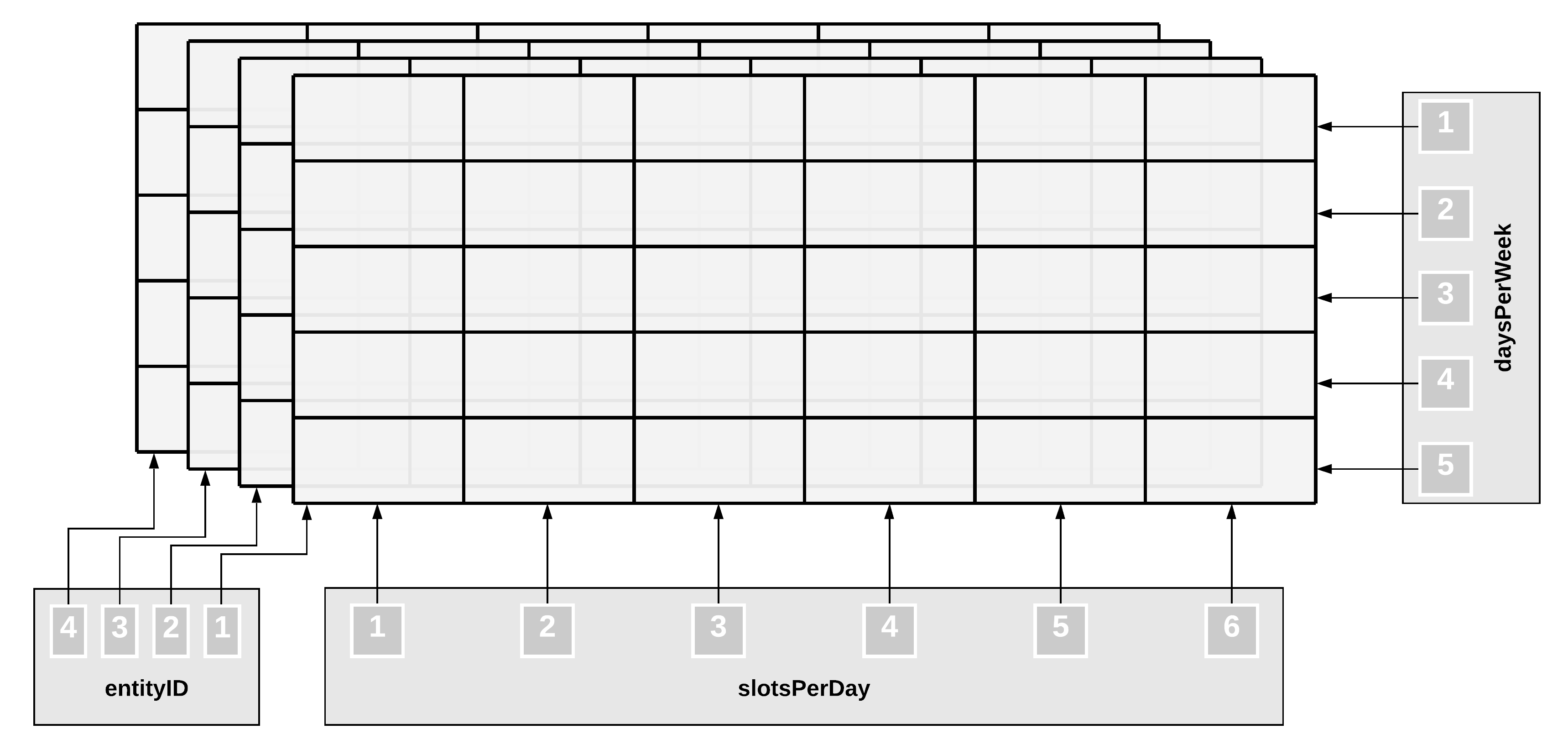}
    \caption{Data-structure of a typical class-routine array where the 3-D space represents the routine of all teachers and sections and each 2-D plane represents the weekly routine of one individual entity}
    \label{fig_routine}
\end{figure}

Our model uses two 3-D arrays as variables to store the routines - one for ``sections" and one for ``teachers", where each 2-D plane represents the routine for a single entity. Fig ~\ref{fig_routine} illustrates this phenomenon.
The model initializes these variables as empty arrays and the solver gradually fills them up.

In this implementation, several parameters, like, the \emph{classDetails}, the \emph{classCredits} parameters, etc, have been used. In \emph{classDetails}, the list of all classes with corresponding teachers and sections is stored. The system allows one, two, or many sections to attend the same class according to need. This can be useful to create sub-sections in an institute. In \emph{classCredits}, the total number of credits, associated with each course in the system, is stored. Another unique feature that this model provides is the use of \emph{teacherBusy} array. With it, the total number of available slots for each enlisted faculty can be reduced.

\subsection{Constraints}
Along with defining all the data-structures, a MiniZinc model must define all possible relations among them. These relations are called constraints. In this paper, we have subdivided all the constraints into two broader classes -- coherence constraints and feature constraints.

\subsubsection{Coherence Constraints}
This type of constraints maintain coherence between the defined variables. In our case, all the routines need to be consistent with each other and the coherence constraints are used to prevent any sort of inconsistency. Some examples are as follows:
\begin{itemize}
    \item \textbf{Individuality Constraint:} A single entity cannot be at two places at the same time. So, for the section routine array and the teacher routine array, constraints had to be declared so that each at a particular time slot, only one class is assigned to an entity.
\begin{lstlisting}[ ]
forall(i in classList,sec in sectionList)(
    if classDetails[i,2] != sec /\
        classDetails[i,3] != sec
        then forall(d in daysPerWeek,s in slotsPerDay)
           (sectionRoutine[sec,d,s] != i) 
        else 
          forall(d in daysPerWeek)
            (sum(s in slotsPerDay)
               (if sectionRoutine[sec,d,s] == i 
                   then 1 
                   else 0 
               endif) <= 1)
    endif);
\end{lstlisting}
\begin{lstlisting}[ ]
forall(i in classList,t in teacherList)(
    if classDetails[i,1] != t
      then forall(d in daysPerWeek,s in slotsPerDay)
        (teacherRoutine[t,d,s] != i) 
      else forall(d in daysPerWeek)
        (sum(s in slotsPerDay)
        (if teacherRoutine[classDetails[i,1],d,s] == i 
            then 1 
            else 0 
        endif) <= 1)
    endif);
\end{lstlisting}
    \item \textbf{Course-entity Constraint:} Sections and the teacher involved in a particular lesson need to have the lesson at the same time slot. More than one constraint was used to ensure this coherence.
\begin{lstlisting}[ ]
forall(t in teacherList,d in daysPerWeek, s in slotsPerDay)(
    if teacherRoutine[t,d,s] > 0
      then 
        sectionRoutine[classDetails[teacherRoutine[t,d,s],2],d,s] == teacherRoutine[t,d,s]  /\ 
        sectionRoutine[classDetails[teacherRoutine[t,d,s],3],d,s] == teacherRoutine[t,d,s]
      else true 
    endif);
\end{lstlisting}
\begin{lstlisting}[ ]
forall(sec in sectionList,d in daysPerWeek, s in slotsPerDay)(
    if sectionRoutine[sec,d,s] > 0
      then teacherRoutine[classDetails[sectionRoutine[sec,d,s],1],d,s] == sectionRoutine[sec,d,s]
      else true 
    endif);
\end{lstlisting}
    \item \textbf{Room Limitation Constraint:} A room cannot host multiple lectures at the same time. So, a constraint was declared that restricted the total number of lectures on a particular slot of a day.
\begin{lstlisting}[ ]
forall(d in daysPerWeek,s in slotsPerDay)(
    sum(t in teacherList)
    (if teacherRoutine[t,d,s]!=0 
        then 1 
        else 0 
    endif) <= rooms);
\end{lstlisting}
\end{itemize}

\subsubsection{Feature Constraints}
The constraints that offer solutions to real-life issues and make our model practically usable are categorized as feature constraints.
\begin{itemize}
    \item \textbf{Routine Generation Constraint:} The most important feature of this system is the generation of routines. It is necessary to ensure every class listed in the \emph{classDetails} array is assigned slots and the teacher and section routines are filled up accordingly. The coherence constraints, in this paper, are designed in a way that this constraint is automatically satisfied.
    \item \textbf{Credit Hours Constraint:} A constraint was declared to make the number of lectures of a particular course equal to the corresponding credit hour provided in \emph{classCredits} array.
\begin{lstlisting}[ ]
forall(i in classList)(
    sum(d in daysPerWeek,s in slotsPerDay)
    (if teacherRoutine[classDetails[i,1],d,s] == i 
        then 1 
        else 0 
    endif) == classCredits[i]);
\end{lstlisting}
    \item \textbf{Teacher Availability Constraint:} Another constraint was declared that prevents assigning slots to a teacher when he/she is busy. The \emph{teacherBusy} array provided by the user is used for this purpose.
\begin{lstlisting}[ ]
forall(t in teacherList, d in daysPerWeek, s in slotsPerDay)(
    if teacherBusy[t,d,s] == 1
    then teacherRoutine[t,d,s] = 0
    else true
    endif);
\end{lstlisting}
    \item \textbf{Non-repeating Lecture Constraint:} The lecture of the same course should not be held multiple times on the same day. This feature is included inside the ``Individuality Constraint".
\end{itemize}

\subsection{Reaching Solution}
\label{sec_solution}
In MiniZinc, the model does not need to explicitly declare the search algorithm to find a satisfiable or optimized solution. Any FlatZinc supported solver can be integrated with the MiniZinc model to obtain the required solution. This paper shows the use of the Chuffed solver \cite{chu2018chuffed} in its model. Chuffed is a lazy clause generation solver \cite{ohrimenko2009propagation}. It combines Finite Domain (FD) propagation \cite{carlsson1997open} with Boolean satisfiability \cite{zhang2002quest}. The FD component of the solver records the reason for each propagation step. Thus an implication graph, similar to the one built by an SAT solver, is generated that creates efficient ``nogoods" and keeps a record of reasons for failure. SAT unit propagation technology uses these ``nogoods" to generate the output. The hybrid solver created in this way has the advantages of both FD solvers and SAT solvers.

\section{Results}
An output of a constraint satisfaction problem always fulfills all the required criteria. Therefore, any generated output of this MiniZinc model is a valid output. So, the performance of the model cannot be evaluated based on the quality of the output. In this paper, the time required to generate a ``satisfiable" solution is used as the metric to evaluate performance.

The total number of courses, the number of teachers, the number of sections - everything can significantly vary for different universities. We conducted experiments with different settings and different data sizes to check which solver performs best and to what extent the model can generate results.

If we consider an average sized university with 4 in-take batches and each batch having 6 courses, then the total number of courses become 24. If each batch has two separable sections and each section has two lab-groups, then the total number of deliverable lectures becomes 48. A system that can successfully generate schedule for such a dataset can be regarded as a successful scheduler. 
In our experiments, our system could successfully generate result for much larger datasets as well. For a dataset with 400 teachers, 400 in-take batches, and 600 courses in-total, our system took around 27 hours to generate the routine on a Ryzen 2600 CPU with 8GB RAM.

\subsection{Solver Comparison}
MiniZinc provides several off-the-shelf solvers. Among them, we tested 8 different solvers \cite{chu2018chuffed,bixby2011gurobi,schulte2006gecode,team2015or,sebastiani2020optimathsat,opturion2016opturion,johnjforrest_2020_3700700,de2013oscar} with our model varying the input parameters. All these are excellent solvers but the results show that the Chuffed solver performs much better than the other solvers for our system.

\begin{table}[htbp]
    \centering
    \caption{Comparing Performance of Different Solvers. The timing (in seconds) of each experiment is denoted in the table. ``Incomplete" results are denoted by `-' sign. The result that took longer than 4 hours to generate is marked with $\infty$ symbol. The results show that Chuffed solver works much better than the result for our model}

    \begin{tabular}{|c|ccccccc|}
        \hline
        Routines & 3  & 6 & 7 & 9 & 16 & 20 & 40  \\
        Teacher(s) & 1  & 2 & 3 & 5 & 8 & 12 & 24  \\
        Section(s) & 2  & 4 & 4 & 4 & 8 & 8 &  16\\
        Courses(s) & 1  & 4 & 6 & 10 & 16 & 24 & 48 \\
        \hline \hline
        Chuffed     & 0.56   & 0.69  & 0.80  & 1.08   & 2.17     & 5.04     & 47.3    \\
        Gecode      & 0.51   & 0.93  & 9362  &$\infty$& $\infty$ & $\infty$ & $\infty$\\
        Optimathsat & -      & -     & -     & -      & -        & -        & -       \\
        Opturion    & -      & -     & -     & -      & -        & -        & -       \\
        Gurobi      & -      & -     & -     & -      & -        & -        & -       \\
        CBC         & 0.88   & 2.6   & 16.7  & 36.9   & 911      & 1680     & $\infty$\\
        OscaR CBLS  & -      & -     & -     & -      & -        & -        & -       \\
        OR Tools    & -      & -     & -     & -      & -        & -        & -       \\
        \hline
    \end{tabular}
    \vspace{0.3 cm}

    \label{tab_solver}
\end{table}
As shown in Table ~\ref{tab_solver}, few of the solvers failed to generate any form of ``complete" result. The process that these solvers use to create search trees and the way our system was designed are not fully compatible. This is the primary reason for this failure. Nevertheless, these solvers are great and redesigning our model can make these solvers generate ``complete" solutions for our model as well.

\subsection{Stress Testing}
Chuffed significantly outperformed the rest of the solvers. Therefore, further experiments were only conducted on it.

\begin{figure}[hbt!]
    \centering
    \includegraphics[width=0.8\linewidth]{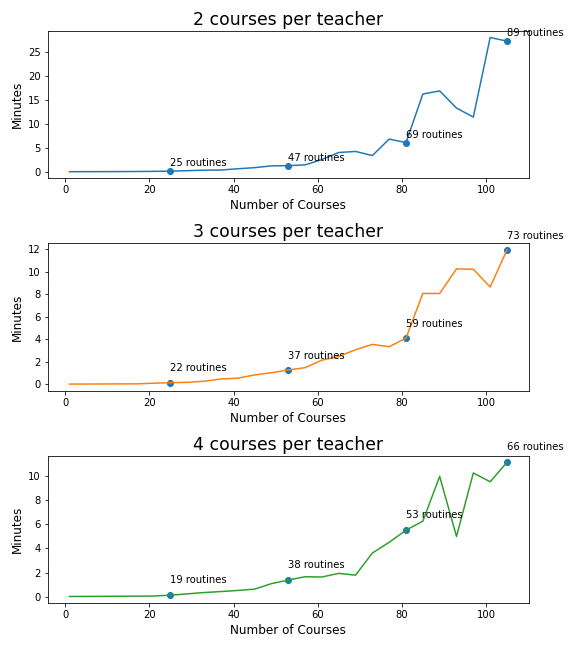}
    \caption{Stress Testing. In Fig ~\ref{fig:stress_test}, timings of data points are plotted against corresponding total number of courses in the system. The number of distinct routines that the model has to generate, are also marked, for certain data points to demonstrate the relation of number of routines and the time required to generate routines. From the stress testing, we see a positive correlation between the number of routines to be generated and required-time to generate those routines.}
    \label{fig:stress_test}
\end{figure}
In the stress testing, data of different sizes and amounts were supplied to the model. Testing was conducted in three phases, each with a different number of courses per teacher. The total number of courses in each testing phase was increased gradually from 2 to 104. In all cases, the model was able to generate satisficing routines for all entities in the system.
 Stress testing was also conducted on datasets of arbitrary values of much larger size as well. 

\section{Discussion}
Our AI-based system can successfully generate class schedule in reasonable time using reasonable computing resources.

\subsection{Constraint Evaluation}
Lewis et al. \cite{lewis2008survey} summarized the use of different meta-heuristics in university time-table scheduling problems. They defined five categories of constraints. These are - Unary constraints, Binary constraints, Capacity constraints, Event-spread constraints, and Agent constraints. Unary constraints deal with single events. For example, a particular class has to be scheduled on a particular day (or a particular slot). In our problem, the way the availability of teachers is handled can be considered a unary constraint. A binary constraint involves multiple events. In our case, the same course cannot be taken on the same day twice. So, this is comparable to the binary constraint. The capacity constraint deals with the capacity of rooms. But in our paper, we have ignored this constraint for simplicity of implementation. The event-spread constraint type is defined as the constraints that spread out the schedule for ease of the students. We do not have any such constraints in our model. The last category of constraints, i.e. the agent constraint, are specific desires by the agent for increased performance. Our system keeps track of teacher's availability which is comparable to the agent constraint.

\subsection{Choice of Solvers}
Any FlatZinc supported solver should, in theory, be able to solve a MiniZinc model if time and memory penalty are not considered. Since different models have different requirements, and time and memory limitations are important factors, all solvers are not always successful in generating output for all models. In the case of the model used in this paper, 8 off-the-shelf solvers were tested and only one of them has produced satisfactory results. 

Out of the 8 solvers, 5 generated ``incomplete" outputs. These solvers are primarily good at solving optimization tasks rather than satisfiability tasks. 2 solvers, Gecode \cite{schulte2006gecode} and CBC \cite{johnjforrest_2020_3700700} can obtain results. But the required time that these solvers take to generate satisficing output grows exponentially with the increase of data size. These solvers do not eliminate sub-trees which is the main reason for this exponential increase in time. For practical usability, the time required for these solvers is too high to be considered. If the total number of courses is brought down to around 5, these solvers can solve in minutes. But for 20+ courses, which is still low for a university, these solvers take few hours to obtain a solution. 

Chuffed \cite{chu2018chuffed}, on the other hand, solves the model in a very quick time. As shown in Section ~\ref{tab_solver}, this solver can solve very large data. The reasons why Chuffed performs better than other solvers are as follows:
\begin{itemize}
    \item Chuffed solver generates ``nogoods" and eliminates a large sub-tree. In class scheduling problems, there are plenty of repetitive sub-trees that will lead to obvious failure. Chuffed can detect them in advance.
    \item Class scheduling task is non-chronological. The use of ``nogoods" allows the solver to make informed choices of back-jumps in such a problem.
    \item Chuffed solver analyzes the conflicts and thus can determine the difference between good and bad decision choices.
\end{itemize}

\section{Conclusion and Future Works}
This paper considers the faculty’s routine and the section’s routine as the primary entities and defines them as two 3-D arrays. Each 2-D plane in these arrays represents the routine of an individual entity (faculty or section). The solver fills up the 3-D arrays with indices of the courses following all the necessary constraints. This AI-driven system relies on the power of MiniZinc modeling language and the efficiency of the Chuffed solver to generate routines in quick time. In less than a minute, the system can generate routines for an average-sized university. Compared to other available systems \cite{havens2010class,caprara2002modeling,socha2002max}, the performance of this university class scheduler is quite remarkable. Not only university but any educational institute following a similar system can also incorporate this system to efficiently generate their class schedule.

The developed system mainly deals with fixed-credit systems. But with few modifications in the architecture, it can be adjusted for open-credit systems as well.

Despite the outstanding performance of the system, the system has plenty of scope for improvements. As the choice of the solver greatly affects the performance of the system, finding the best possible solver can be the primary goal of the future works. As shown in Sec~\ref{tab_solver}, all experiments was conducted with off-the-shelf solvers. But in the future, we plan to build a custom solver that will further improve the performance of the system. Since Chuffed has the best comparative performance, the custom solver can be built based on it.

\section*{Acknowledgements}
The research was funded by University Grant Commission (UGC) Bangladesh. 

\bibliographystyle{bib_style.bst}
\bibliography{my_bib.bib}
\vspace{12pt}

\end{document}